\documentclass[final]{cvpr}

\usepackage{times}
\usepackage{epsfig}
\usepackage{graphicx}
\usepackage{amsmath}
\usepackage{amssymb}

\usepackage{hhline}
\usepackage{xcolor}
\usepackage{colortbl}
\usepackage{floatrow}
\usepackage{subcaption}

\usepackage[normalem]{ulem}

\usepackage{siunitx}
\usepackage{acronym}

\acrodef{cnn}[CNN]{convolutional neural network}
\acrodef{dnn}[DNNs]{deep neural networks}
\acrodef{3d}[3D]{three-dimensional}
\acrodef{2d}[2D]{two-dimensional}
\acrodef{vo}[VO]{visual odometry}
\acrodef{slam}[SLAM]{simultanious localozation and mapping}
\acrodef{mvs}[MVS]{multi-view stereo}
\acrodef{6dof}[6DoF]{6 degrees of freedom}
\acrodef{ssim}[SSIM]{structural similarity index measure}
\acrodef{sad}[SAD]{sum of absolute differences}

\acused{6dof}
\acused{slam}
\acused{3d}
\acused{2d}

\usepackage[pagebackref=true,breaklinks=true,colorlinks,bookmarks=false]{hyperref}

\definecolor{drawioBlue}{HTML}{DAE8FC}
\definecolor{drawioOrange}{HTML}{FFE6CC}
\definecolor{drawioGreen}{HTML}{D5E8D4}

\newcommand\Tstrut{\rule{0pt}{2.6ex}}         %
\newcommand\Bstrut{\rule[-0.9ex]{0pt}{0pt}}   %

\begin{document}

\title{MonoRec: Semi-Supervised Dense Reconstruction in Dynamic
	Environments from a Single Moving Camera}

\author{Felix Wimbauer$^{1,\star}$\quad  Nan Yang$^{1, 2, \star}$ \quad Lukas von Stumberg$^{1}$ \quad Niclas Zeller$^{1, 2}$ \quad Daniel Cremers$^{1, 2}$ \\$^1$ Technical University of Munich, $^2$ Artisense \\
{\tt\small \{wimbauer, yangn, stumberg, zellern, cremers\}@in.tum.de}
}

\maketitle

\begin{abstract}
	In this paper, we propose MonoRec, a semi-supervised monocular dense
	reconstruction architecture that predicts depth maps from a single moving
	camera in dynamic environments. MonoRec
	is based on a \acl*{mvs} setting which encodes the information of
	multiple consecutive images in a cost volume. To deal with dynamic objects
	in the scene, we introduce a MaskModule that predicts moving object masks
	by leveraging the photometric inconsistencies encoded in the cost
	volumes.
	Unlike other \acl*{mvs} methods, MonoRec is
	able to reconstruct both static and moving objects by 
	leveraging
	the predicted masks.
	Furthermore, we present a novel multi-stage training
	scheme with a semi-supervised loss formulation that does not require LiDAR depth values.
	We carefully evaluate MonoRec on the KITTI dataset and show that it
	achieves state-of-the-art performance compared to both multi-view and
	single-view methods. With the model trained on KITTI, we furthermore demonstrate
	that MonoRec is able to generalize well to both the Oxford RobotCar dataset and the
	more challenging TUM-Mono dataset recorded by a handheld camera. Code and 
	related materials are available at 
	\url{https://vision.in.tum.de/research/monorec}.
\end{abstract}

\let\thefootnote\relax\footnotetext{$\star$ Indicates equal contribution.}

\begin{figure}
	\begin{center}
		\includegraphics[width=\textwidth]{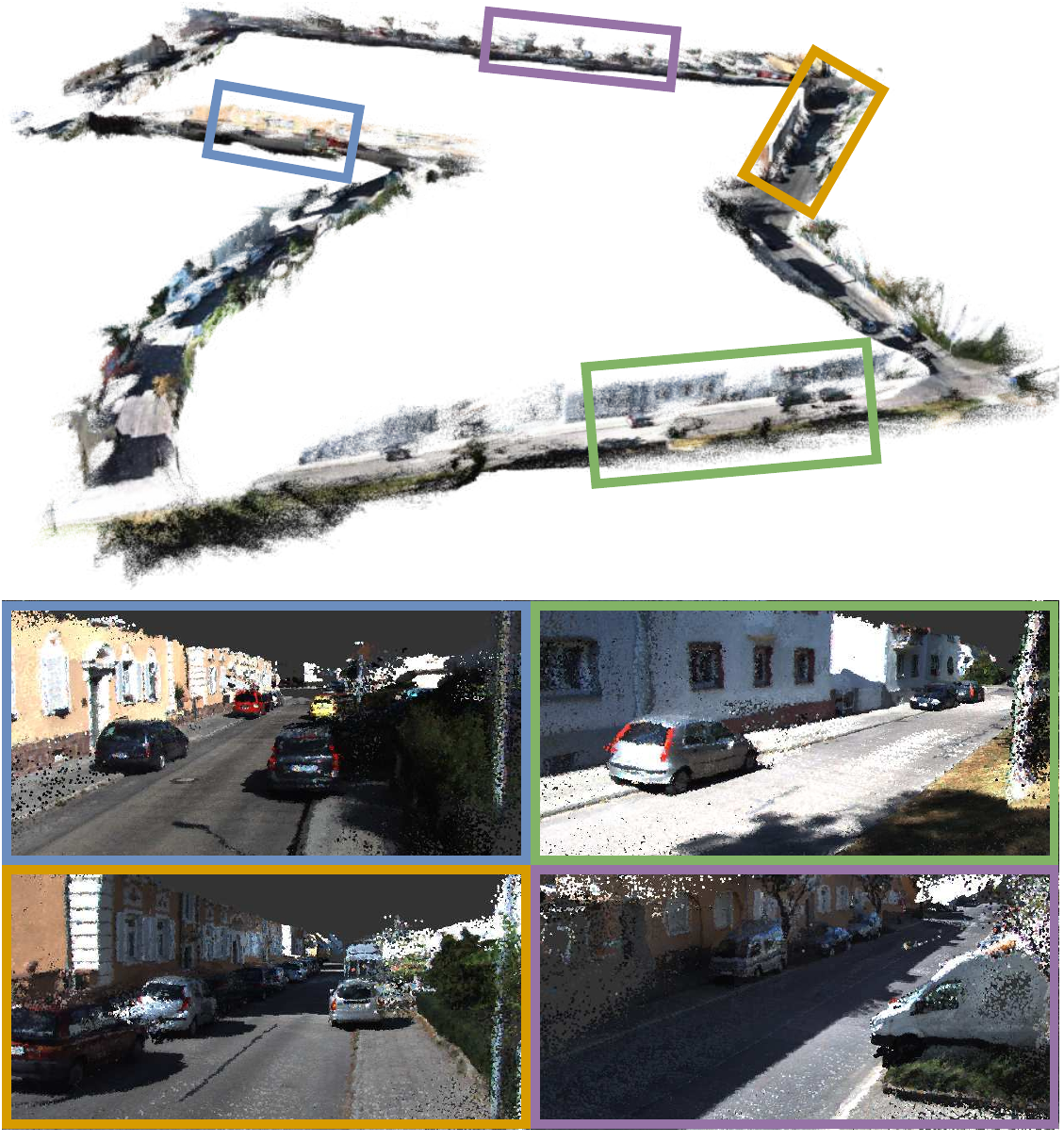}\\[-0.5cm]
		\caption{MonoRec can deliver high-quality dense reconstruction from a 
		single moving camera. The figure shows an example of a large-scale 
		outdoor point cloud reconstruction (KITTI Odometry sequence 07) by 
		simply accumulating predicted depth maps. Please refer to our project 
		page for the video of the entire reconstruction of the sequence.
		}
		\label{fig:teaser}
	\end{center}
\end{figure}

\section{Introduction}
\label{sec:introduction}
\subsection{Real-world Scene Capture from Video}
Obtaining a \ac{3d} understanding of the entire static and dynamic environment 
can be seen as one of the key-challenges in robotics, AR/VR, and autonomous 
driving.
State of today, this is achieved based on the fusion of multiple sensor sources (incl. cameras, LiDARs, RADARs and IMUs).
This guarantees dense coverage of the vehicle's surroundings and accurate ego-motion estimation.
However, driven by the high cost as well as the challenge to maintain 
cross-calibration of such a complex sensor suite, there is an increasing demand 
of reducing the total number of sensors. Over the past years, researchers have therefore 
put a lot of effort into solving the problem of perception with only a single 
monocular camera.
Considering recent achievements in monocular \ac{vo} \cite{Engel2018dso, Yang2020d3vo, Usenko2020}, with respect to ego-motion estimation, this was certainly successful.
Nevertheless, reliable dense \ac{3d} mapping of the static environment and moving objects is still an open research topic.

To tackle the problem of dense \ac{3d} reconstruction based on a single moving 
camera, there are basically two parallel lines of research.
On one side, there are dense \ac{mvs} methods, which evolved over the 
last decade \cite{Pizzoli2014, Schoenberger2016, Campbell2008} and saw a great 
improvement through the use of \acp{cnn} \cite{huang2018deepmvs, Yao2018, Yang2020}.
On the other side, there are monocular depth prediction methods which purely 
rely on deep learning \cite{Eigen2014, Godard2017, Yang2020d3vo}.
Though all these methods show impressive performance, both types have also their 
respective shortcomings.
For \ac{mvs} the overall assumption is a stationary environment to be 
reconstructed, so the presence of dynamic objects deteriorate their 
performance.
Monocular depth prediction methods, in contrast, perform very well in 
reconstructing moving objects, as predictions are made only based on individual 
images. At the same time, due to their use of a single image only, they strongly 
rely on the perspective appearance of objects as observed with specific camera 
intrinsics and extrinsics and therefore do not generalize well to other 
datasets.

\subsection{Contribution}
To combine the advantage of both deep \ac{mvs} and monocular depth prediction, 
we 
propose MonoRec, a novel monocular dense reconstruction architecture that 
consists of a MaskModule and a DepthModule. We encode the information from 
multiple consecutive images using cost volumes which are constructed based on 
\ac{ssim}~\cite{wang2004image} instead of \ac{sad} like prior works. The 
MaskModule is able to 
identify moving pixels and downweights the corresponding voxels in the cost 
volume.
Thereby, in contrast to other \ac{mvs} methods, MonoRec does not suffer from artifacts on moving 
objects and therefore delivers depth estimations on both static and dynamic objects.

With the proposed multi-stage training scheme, MonoRec achieves 
state-of-the-art performance compared to other \ac{mvs} and monocular depth 
prediction methods on the KITTI dataset \cite{Geiger2013kitti}.
Furthermore, we validate the generalization capabilities of our network on the 
Oxford RobotCar dataset \cite{Maddern2017robotcar} and the TUM-Mono 
dataset~\cite{engel2016monodataset}. 
Figure~\ref{fig:teaser} 
shows a dense point cloud reconstructed by our method on one of our test 
sequences of KITTI.

\section{Related Work}
\label{sec:related_work}

\subsection{Multi-view Stereo}
Multi-view stereo (\acs{mvs}) methods estimate a dense representation of the 
\ac{3d} environment based on a set of images with known poses.
Over the past years, several methods have been developed to solve 
the \ac{mvs} problem \cite{Seitza1997, Kutulakos1999, Lhuillier2005, 
Campbell2008, Stuehmer2010, Tola2011, Pizzoli2014, Galliani2015, 
Schoenberger2016, Yao2017} based on classical optimization.
Recently, due to the advance of \ac{dnn}, different learning based approaches 
were proposed.
This representation can be volumetric \cite{Ji2017, Kar2017, Murez2020} or 
\ac{3d} point cloud based \cite{Chen2019, Furukawa2010}.
Most popular are still depth map representations predicted from a 
\ac{3d} cost volume \cite{huang2018deepmvs, Wang2018, Yao2018, zhou2018deeptam, 
Hou2019, Xue2019, Romanoni2019, Im2019, Luo2019, Yao2019, Gu2020, Yu2020, 
Yang2020}.
Huang et al. \cite{huang2018deepmvs} proposed one of the first cost-volume based approaches.
They compute a set of image-pair-wise plane-sweep volumes with respect to a 
reference image and use a \ac{cnn} to predict one single depth map based on 
this set. Zhou et al. \cite{zhou2018deeptam} also use the photometric cost 
volumes as the inputs of the deep neural networks and employ a two stage 
approach for dense depth prediction.
Yao et al. \cite{Yao2018} instead calculate a single 
cost volume using deep features of all input images.

\subsection{Dense Depth Estimation in Dynamic Scenes}
Reconstructing dynamic scenes is 
challenging since the moving objects violate the static-world assumption for 
classical multi-view stereo methods. Russell et al.~\cite{russell2014video} and 
Ranftl et al.~\cite{ranftl2016dense} base on motion segmentation and perform 
classical optimization. Li et al.~\cite{li2019learning} proposed to 
estimate dense depth maps from the scenes with moving people. All these methods 
need additional inputs, e.g., optical flow, object masks, etc., for the 
inference, while MonoRec requires only the posed images as the inputs. Another 
line of research is monocular depth 
estimation~\cite{Eigen2014, 
eigen2015predicting, Laina2016, Li2015, fu2018deep, yang2018deep, Godard2017, 
	Tateno2017, Zhou2017, Yin2018, Zhan2018, Wang2018a, Gordon2019, 
	godard2019digging, Yang2020d3vo}. These methods are not affected by moving 
objects, but the 
depth estimation is not necessarily accurate, especially in unseen scenarios. 
Luo et al.~\cite{Luo-VideoDepth-2020} proposed a test-time 
optimization method which is not real-time capable. In a concurrent work, Watson 
et al.~\cite{watson2021temporal} address moving objects with the consistency 
between monocular depth estimation and multi-view stereo, while MonoRec 
predicts the dynamic masks explicitly by the proposed MaskModule.

\begin{figure*}
	\begin{center}
		\includegraphics[width=\textwidth]{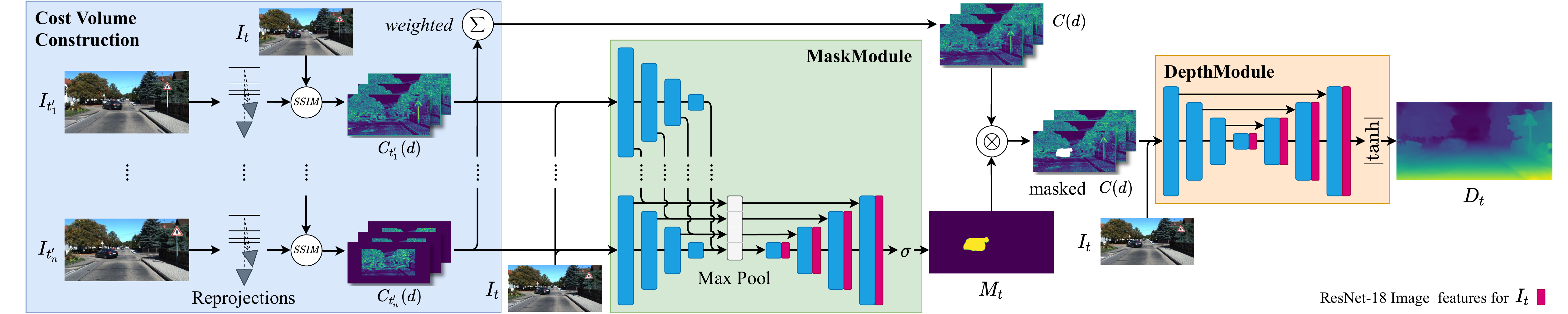}\\[-0.5cm]
		\caption{\textbf{MonoRec Architecture}: It first 
			constructs a 
			photometric cost volume from multiple input frames. Unlike prior 
			works, we use the \ac{ssim}~\cite{wang2004image} metric 
			instead of \ac{sad} to
			measure the photometric consistency. The MaskModule
			aims to detect 
			inconsistencies between the different input frames to determine 
			moving objects. 
			The multi-frame cost volume $C$ is multiplied with the predicted 
			mask and then passed 
			to the DepthModule which predicts a dense inverse 
			depth map. In both the decoders of MaskModule and DepthModule, the 
			cost 
			volume features are concatenated with pre-trained ResNet-18 
			features.}
		\label{fig:arch}
	\end{center}
\end{figure*}

\subsection{Dense SLAM}
Several of the methods cited above solve both the problem of dense \ac{3d} 
reconstruction and camera pose estimation~\cite{Tateno2017, Zhou2017, Yin2018, Zhan2018, 
zhou2018deeptam, yang2018deep, Yang2020d3vo}.
Nevertheless, these methods either solve both problems independently or only integrate one 
into the other (e.g. \cite{zhou2018deeptam, Yang2020d3vo}). Newcombe et al. 
\cite{Newcombe2011} instead jointly optimize the \ac{6dof} camera pose and the 
dense \ac{3d} scene structure.
However, due to its volumetric map representation it is only applicable to small-scale scenes.
Recently, Bloesch et al. \cite{Bloesch2018} proposed a learned code 
representation which can be optimized jointly with the \ac{6dof} camera 
poses. This idea is pursued by Czarnowski et al. \cite{Czarnowski2020} and 
integrated into a full \ac{slam} system. All the above-mentioned methods, 
however, do not address the issue 
of moving objects.
Instead, the proposed MonoRec network explicitly deals with moving objects and 
achieves superior 
accuracy both on moving and on static structures. Furthermore, prior works show 
that the accuracy of camera tracking does not necessarily improve with more 
points~\cite{Engel2018dso,fontan2020information}. MonoRec therefore focuses 
solely on delivering dense reconstruction using poses from a sparse VO 
system and shows state-of-the-art results on public benchmarks. Note that, this 
way, MonoRec can be easily combined with any VO systems with 
arbitrary sensor setups.

\section{The MonoRec Network}
\label{sec:method}

MonoRec uses a set of consecutive frames and the corresponding camera poses to 
predict a dense depth map for the given keyframe.
The MonoRec architecture combines a MaskModule and a DepthModule.
MaskModule predicts moving object masks that improve depth accuracy and allows us
to eliminate noise in \ac{3d} reconstructions.
DepthModule predicts a depth map from the masked cost volume.
In this section, we first describe 
the different modules of our architecture, and then discuss the specialized 
multi-stage semi-supervised training scheme.

\subsection{Preliminaries}

Our method aims to predict a dense inverse depth map $D_t$ of the 
selected keyframe from a set of consecutive frames 
$\{I_1,\cdots,I_N\}$. We denote the selected keyframe as $I_t$ 
and others as $I_{t'}$ ($t'\in\{1,\cdots,N\} \setminus t$). Given the camera 
intrinsics, the inverse 
depth map $D_t$, and the 
relative camera pose $\mathbf{T}_{t'}^t \in \mathrm{SE}(3)$ between $I_{t'}$ and 
$I_t$, we can perform the reprojection from $I_{t'}$ to $I_t$ as
\begin{equation}
I_{t'}^{t} = I_{t'}\left< \mathit{proj}\left(D_{t}, 
\mathbf{T}_{t'}^t\right)\right>,
\label{eq:image_warping}
\end{equation}
where $\mathit{proj()}$ is the projection function and $\left< \right>$ is the 
differentiable sampler~\cite{Jaderberg2015}. This reprojection 
formulation is important for both the cost volume formation (Sec.~\ref{sec:cv}) 
and the self-supervised loss term (Sec.~\ref{sec:ms-t}).

In the following, we refer to the consecutive frames as temporal
stereo (\textbf{T}) frames. During training, we use an additional static stereo (\textbf{S})
frame $I_{t^S}$ for each sample, which was captured by a synchronized stereo camera at the same time as the respective keyframe.

\subsection{Cost Volume}
\label{sec:cv}

A cost volume encodes geometric information from the different frames 
in a tensor that is suited as input for neural networks. 
For a number of discrete depth steps, the temporal stereo frames are reprojected to 
the keyframe and a pixel-wise photometric error is computed. Ideally, the 
lower the photometric error, the better the depth step approximates the real 
depth at a given pixel.
Our cost volume follows the general formulation of the prior works~\cite{Newcombe2011,zhou2018deeptam}. 
Nevertheless, unlike the previous works that define the photometric error 
$pe()$ as a patch-wise \ac{sad}, we propose to use the \ac{ssim} as follows:
\begin{equation}
pe(\mathbf{x}, d) = \frac{1 - 
	\text{SSIM}(I_{t'}^{t}(\mathbf{x}, d), I_t(\mathbf{x}))}{2}
\end{equation}
with $3\times3$ patch size.
Here $I_{t'}^{t}(\mathbf{x}, d)$ defines the intensity at pixel $\mathbf{x}$ of the image $I_{t'}$ 
warped with constant depth $d$.
In practice, we clamp the error to $[0,1]$. 
The cost 
volume $C$ stores at $C(\mathbf{x}, d)$ the aggregated photometric consistency for pixel 
$\mathbf{x}$ and 
depth $d$
\begin{equation}
C(\mathbf{x}, d) = 1 - 2 \cdot \frac{1}{\sum_{t'} \omega_{t'}} \cdot 
\sum_{t'} pe_{t'}^{t} (\mathbf{x}, d)\cdot\omega_{t'}(\mathbf{x})
\end{equation}
where $d \in \{d_i | d_{min} + \frac{i}{M} \cdot (d_{min} 
-d_{max})\}$.
The 
weighting term $w_{t'}(\mathbf{x})$ weights the 
optimal depth step height based on the photometric error while others are weighted lower:
\begin{equation}
\begin{split}
w_{t'}(\mathbf{x}) =& 1 - \frac{1}{M - 1} \\ \cdot &\sum_{d \neq d^*}  \exp 
\left(-\alpha \left(pe_{t'}^{t}(\mathbf{x},d) - 
pe_{t'}^{t}(\mathbf{x},d^*)\right)^2\right)
\end{split}
\end{equation}
with $d^*_{t'} = \arg \min_d 
pe_{t'}^{t}(\mathbf{x},d)$. Note that $C(\mathbf{x}, d)$ has the 
range $[-1,1]$ where $-1/1$ indicates the lowest/highest photometric 
consistency.

In the following section, we denote cost volumes calculated based on the 
keyframe $I_t$ and only \textit{one} non-keyframe $I_{t'}$ by 
$C_{t'}(\mathbf{x},d)$ 
where applicable.

\subsection{Network Architecture}
As shown in Figure~\ref{fig:arch}, the proposed network architecture contains 
two sub-modules, namely, MaskModule and DepthModule.

\paragraph{MaskModule}
MaskModule aims to 
predict a mask $M_t$ where 
$M_t(\mathbf{x}) \in [0,1]$ indicates the probability of a pixel $\mathbf{x}$ in $I_t$ belonging to a moving 
object. Determining moving objects from $I_t$ alone is an 
ambiguous task and hard to be generalizable. Therefore, we 
propose to use the set of cost volumes $\{C_{t'} | t' \in \{1,\cdots,N\} 
\setminus 
t\}$ 
which encode the geometric priors between $I_t$ and $\{I_{t'} | t' \in \{1,\cdots,N\} 
\setminus t\}$ respectively. We use 
$C_{t'}$ instead of $C$ since the 
inconsistent geometric information from different $C_{t'}$ is a strong prior 
for 
moving object prediction -- dynamic pixels yield 
inconsistent optimal depth steps in different $C_{t'}$.
However, geometric priors alone are not enough to predict moving objects, since 
poorly-textured or non-Lambertian surfaces can lead to inconsistencies as well. 
Furthermore, the cost volumes tend to reach a consensus on wrong depths that semantically don't fit into the context of the scene for objects that move at constant speed
.
Therefore, we further leverage pre-trained 
ResNet-18~\cite{he2016deep} features of $I_t$
to encode semantic priors in addition to the geometric ones.
The network adapts a U-Net 
architecture design \cite{ronneberger2015u} with skip connections. All cost volumes are passed through the encoders with shared weights. The features 
from different cost volumes are 
aggregated using max-pooling and then passed through the decoder. 
In this way, MaskModule can be applied to different 
numbers of frames without retraining.

\paragraph{DepthModule} DepthModule predicts a
dense pixel-wise inverse depth map $D_{t}$ of $I_t$. To this end, the module 
receives the complete cost volume $C$ concatenated with the keyframe $I_t$. 
Unlike MaskModule, here we 
use $C$ instead of $C_{t'}$ since multi-frame cost volumes in general 
lead to higher depth accuracy and robustness against 
photometric noise~\cite{Newcombe2011}.
To eliminate wrong depth predictions for moving objects, we perform pixel-wise
multiplication between $M_t$ and the cost volume $C$ for every 
depth step $d$.
This way, there won't be any maxima (\ie strong priors) in regions of moving objects left, 
such that DepthModule has to rely on information from the image features 
and the surroundings to infer the depth of moving objects. We employ a 
U-Net architecture with multi-scale depth outputs from the 
decoder~\cite{godard2019digging}.
Finally, DepthModule outputs an 
interpolation factor between $d_{min}$ and $d_{max}$. In practice, we use $s=4$ 
scales of depth prediction.

\begin{figure}
	\begin{center}
		\includegraphics[width=\textwidth]{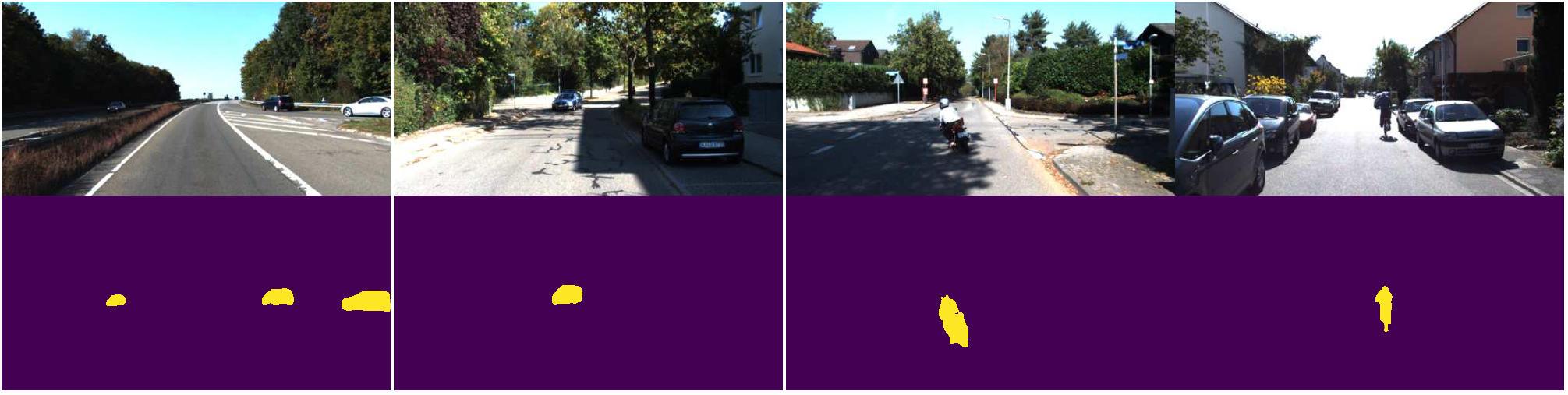}\\[-0.5cm]
		\caption{\textbf{Auxiliary Training Masks}: Examples of auxiliary 
		training masks from the training set that are used as reference.}
		\label{fig:gt_masks}
	\end{center}
\end{figure}

\subsection{Multi-stage Training}
\label{sec:ms-t}

In this section, we propose a multi-stage training scheme for the 
networks. Specifically, the bootstrapping stage, the MaskModule refinement 
stage 
and the DepthModule refinement stage are executed successively.

\paragraph{Bootstrapping}
\label{sec:pre-t}
In the bootstrapping stage, MaskModule and DepthModule are 
trained 
separately. DepthModule takes the \textit{non-masked} $C$ as the input and 
predicts $D_t$. The training objective of DepthModule is defined as a 
multi-scale ($s\in[0, 3]$) semi-supervised loss.
It combines a self-supervised 
photometric loss and an edge-aware smoothness term, as proposed in 
\cite{godard2019digging}, with a supervised sparse depth loss.
\begin{equation}
\mathcal{L}_{depth} = \sum_{s=0}^{3} \mathcal{L}_{self, s} + \alpha 
\mathcal{L}_{sparse, s} + \beta 
\mathcal{L}_{smooth, s}.
\end{equation}
The self-supervised loss is computed from the photometric errors between the 
keyframe and the reprojected temporal stereo and static stereo frames:
\begin{equation}
\begin{split}
	\mathcal{L}_{self, s} &= \min_{t^\star \in t' \cup \{t^S\}}\biggl(\lambda\frac{1 - 
		\text{SSIM}(I_{t^\star}^{t}, I_t)}{2} \\ &\qquad\qquad\qquad\qquad+ 
		(1-\lambda)||I_{t^\star}^{t} 
		- 
		I_t||_1\biggr) ,
\end{split}
\end{equation}
where $\lambda = 0.85$. Note that $\mathcal{L}_{self, s}$ takes the per-pixel 
minimum which has be shown to be superior compared to the per-pixel 
average~\cite{godard2019digging}.
The sparse supervised depth
loss is defined as
\begin{equation}
\mathcal{L}_{sparse, s} = ||D_t - D_{VO}||_1,
\end{equation}
where the ground-truth sparse depth maps ($D_{VO}$) are obtained by a visual odometry system~\cite{yang2018deep}. Note 
that all the supervision signals of DepthModule are generated from either 
images themselves or the visual odometry system without 
any manual labeling or LiDAR depth.

MaskModule is trained with the mask loss $\mathcal{L}_{mask}$ which 
is the weighted binary cross entropy between the predicted mask $M_t$ and the auxiliary 
ground-truth moving object mask $M_{aux}$. We generate $M_{aux}$ by leveraging 
a pre-trained Mask-RCNN and the trained DepthModule as explained above. We 
firstly define the movable object classes, e.g., cars, cyclists, etc, and 
then obtain the instance segmentations of these object classes for the training 
images. A \textit{movable} instance is classified as a \textit{moving} instance 
if it 
has a 
high ratio of photometrically inconsistent pixels between temporal stereo and 
static stereo. Specifically, for each image, we predict its depth maps 
$D_t$ and $D_t^{S}$ using the cost volumes formed by temporal stereo images $C$
and static stereo images $C^S$, respectively. Then a pixel $\mathbf{x}$ 
is regarded as a moving pixel if two of 
the following three metrics are above predefined thresholds: (1) The static stereo photometric 
error using $D_t$, \ie, 
$pe_{t^S}^t (\mathbf{x}, D_t(\mathbf{x}))$. (2) The 
average temporal stereo photometric error using $D_t^{S}$, \ie, 
$\overline{pe_{t'}^t} (\mathbf{x}, 
D_t^{S}(\mathbf{x}))$. 
(3) The difference between $D_t(\mathbf{x})$ and 
$D_t^{S}(\mathbf{x})$.
Please 
refer to our supplementary materials for more details.  
Figure~\ref{fig:gt_masks} shows some examples of the generated auxiliary 
ground-truth moving object masks.

\begin{figure}
	\begin{center}
		\includegraphics[width=\textwidth]{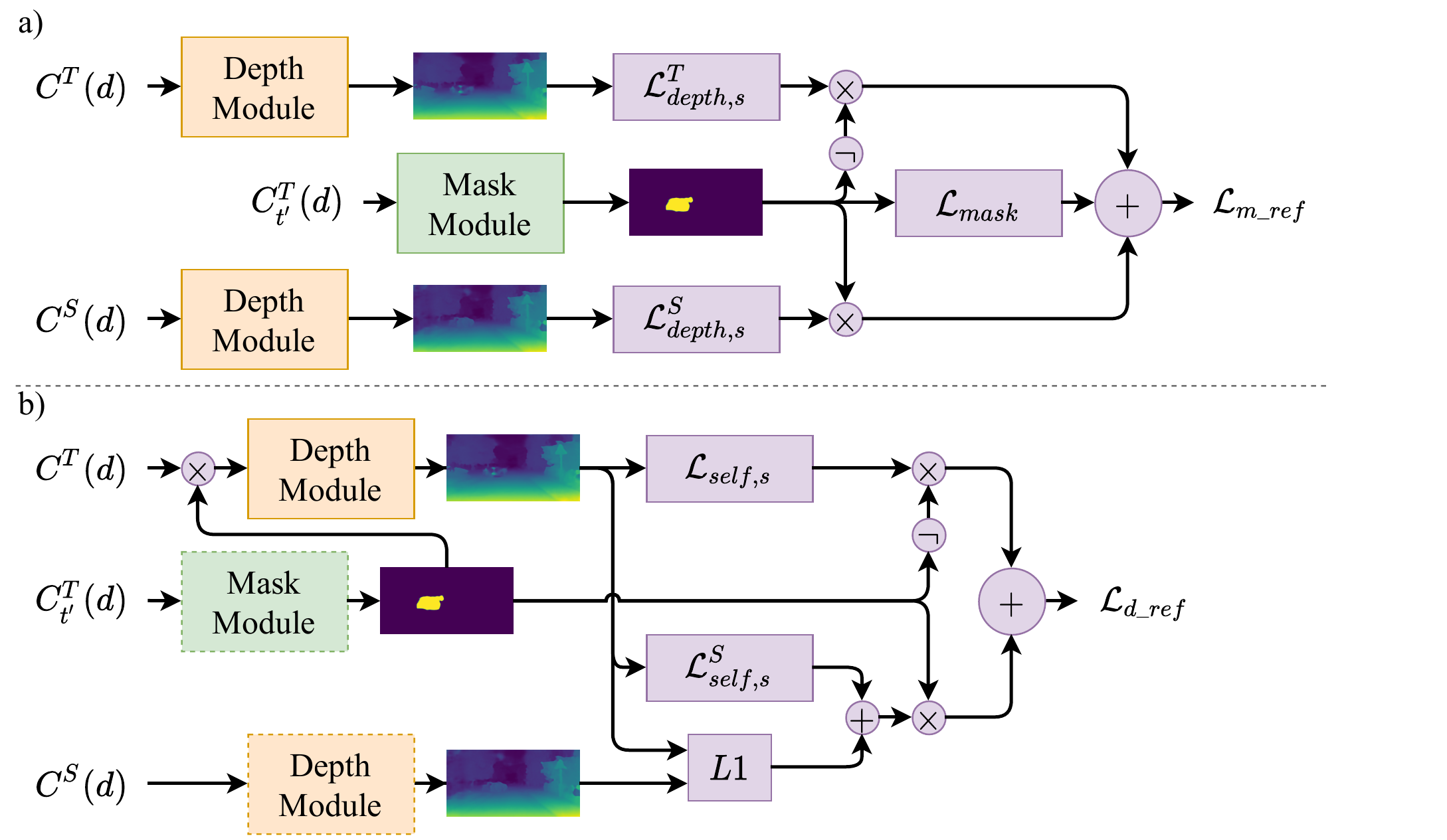}\\[-0.5cm]
		\caption{\textbf{Refinement Losses}: a) 
		MaskModule refinement and b) DepthModule refinement loss 
		functions. 
		Dashed outlines denote that no gradient is being computed for the 
		respective forward pass in the module.}
	\label{fig:refine}
	\end{center}
\end{figure}

\paragraph{MaskModule Refinement}
\label{sec:att-ref}
The bootstrapping stage for MaskModule is limited in two ways: (1) 
Heavy augmentation is needed since mostly only a very small percentage 
of pixels on the image belongs to moving objects. (2) The auxiliary masks are 
not necessarily related to the geometric prior in the cost volume, 
which slows down the convergence. Therefore, to improve the mask prediction, we 
utilize the 
trained DepthModule from the bootstrapping stage. We leverage the fact that 
the depth prediction for moving objects, and 
consequently the 
photometric consistency, should be better with a static stereo prediction than with
a temporal stereo one. Therefore, similar to the classification of moving pixels as 
explained in the previous section, we obtain $D_t^{S}$ and $D_t$ from two 
forward passes using $C^S$ and $C$ as inputs, respectively. Then we compute the 
static stereo photometric error $L_{self, s}^{\prime S}$ using $D_t^S$ as depth and the temporal 
stereo photometric error $L_{self, s}^{\prime T}$ using $D_t$ as depth. To 
train $M_{t}$, we interpret it as pixel-wise 
interpolation factors between $\mathcal{L}_{self, s}^{\prime S}$ and 
$\mathcal{L}_{self, s}^{\prime T}$, and minimize the summation:
\begin{equation}
\begin{split}
\mathcal{L}_{m\_ref} =
    & \sum_{s=0}^{3}\left( M_{t}\mathcal{L}_{depth, 
	s}^{\prime S} + (1 - M_{t}) \mathcal{L}_{depth, 
	s}^{\prime T}\right) \\
    & +\mathcal{L}_{mask}.
\end{split}
\end{equation}
Figure~\ref{fig:refine}(a) shows the diagram illustrating different loss terms. 
Note that we still add the supervised mask loss 
$\mathcal{L}_{mask}$ as a 
regularizer to stabilize the training. This way, the new gradients are directly 
related to the geometric structure in the cost volume and help to improve the 
mask prediction accuracy and alleviate the danger of overfitting.

\paragraph{DepthModule Refinement}
\label{sec:depth-ref}
The bootstrapping stage does not distinguish between 
the moving pixels and static pixels when training DepthModule. Therefore, we 
aim to refine DepthModule such that it is able to predict proper depths also 
for moving objects.
The key idea is that, by utilizing $M_{t}$, only 
the static stereo loss is backpropagated for moving pixels, while for static pixels the temporal stereo, static stereo and sparse depth losses are backpropagated. 
Because moving objects make up only a small percentage of all pixels in a 
keyframe, the gradients from the photometric error are rather weak. To solve 
this, we perform a further static stereo forward pass and use the resulting 
depth map $D_t^{S}$ as prior for moving objects.
Therefore, as shown in Figure~\ref{fig:refine}(b), the loss for refining 
DepthModule is defined as 
\begin{equation}
\begin{split}
\mathcal{L}_{d\_ref, s} =
&(1 - M_{t}) \left(\mathcal{L}_{self, s} + \alpha 
\mathcal{L}_{sparse, s}\right) \\ 
& + M_{t} \left(\mathcal{L}_{self, s}^S + \gamma \left|\left|D_t - D_t^S\right|\right|_1 \right) \\
& + \beta \mathcal{L}_{smooth, s}.
\end{split}
\end{equation}

\subsubsection{Implementation Details}
The networks are implemented in PyTorch~\cite{paszke2019pytorch} with image 
size $512 \times 256$. For the bootstrapping stage, we train DepthModule for 
70 
epochs with learning rate 
$lr = 1e ^{-4}$ for the first 65 epochs and $lr = 1e^{-5}$ for the 
remaining ones. MaskModule is trained for 60 epochs with $lr = 
1e^{-4}$. During MaskModule refinement, we train for 32 epochs with 
$lr = 
1e^{-4}$, and during DepthModule refinement we train for 15 epochs with 
$lr = 
1e^{-4}$ and another 4 epochs at $lr = 1e^{-5}$. The hyperparameters $\alpha$, 
$\beta$ and $\gamma$ are set to $4$, $10^{-3} \times 
2^{-s}$ and $4$, respectively.
For inference, MonoRec can achieve 10 fps with batch size 1 using 2GB 
memory.

\section{Experiments}
\label{sec:experiments}

\begin{figure*}
	\begin{center}
		\includegraphics[width=1.\textwidth]{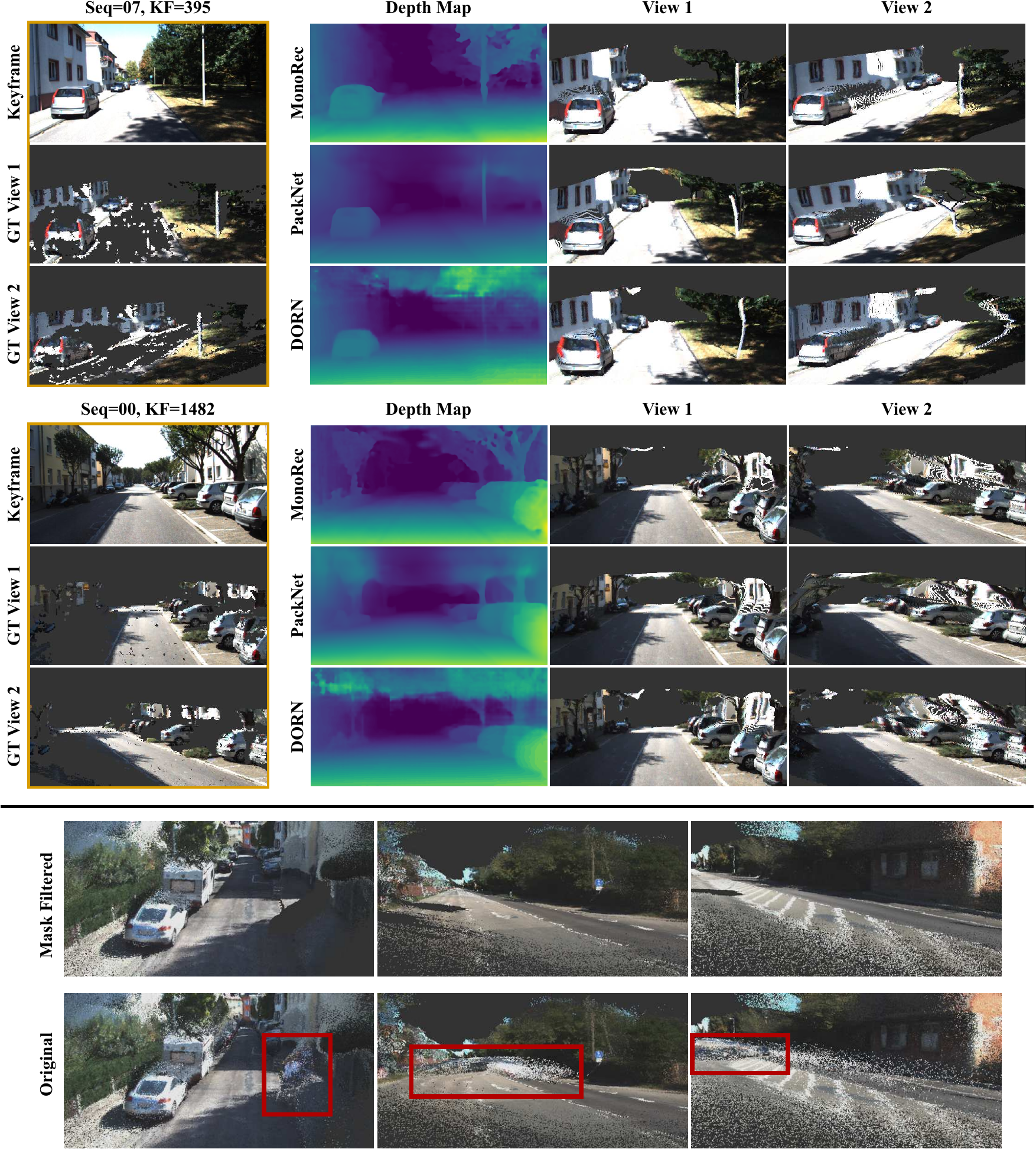}
		\caption{\textbf{Qualitative Results on KITTI}: The upper part of the 
		figure shows the results for a selected 
		number of frames from the KITTI test set. The compared PackNet model 
		was trained in a 
		semi-supervised fashion using LiDAR as the ground truth. Besides the 
		depth maps, we also show the 3D point clouds by reprojecting the depth 
		and viewing from two different perspectives. For comparison we show the 
		LiDAR ground truth from the corresponding perspectives. Our method 
		clearly shows the best prediction quality. The lower part of the figure 
		shows large scale reconstructions as point clouds accumulated from 
		multiple frames. The red insets depict the reconstructed artifacts 
		from moving objects. With the proposed MaskModule, we can 
		effectively filter out the moving objects to avoid those artifacts 
		in the final reconstruction.}
		\label{fig:qualitative_results}
		\end{center}
\end{figure*}

\begin{table*}
	\begin{center}
		\scriptsize
		\begin{tabular}{|l|ccc|cccc|ccc|}
			\hline
			Method & Training & Dataset & Input & \cellcolor{drawioOrange}Abs 
			Rel & \cellcolor{drawioOrange}Sq Rel & \cellcolor{drawioOrange}RMSE 
			& \cellcolor{drawioOrange}RMSE$_{log}$ & 
			\cellcolor{drawioGreen}$\delta<1.25$ & 
			\cellcolor{drawioGreen}$\delta<1.25^2$ & 
			\cellcolor{drawioGreen}$\delta<1.25^3$ \Tstrut\\
			\hhline{===========}
			Colmap \cite{schonberger2016structure} (geometric) & - & - & KF + 2 & 0.099 & 3.451 & 5.632 & 0.184 & 0.952 & 0.979 & 0.986 \Tstrut\\
			Colmap \cite{schonberger2016structure} (photometric) & - & - & KF + 2 & 0.190 & 6.826 & 7.781 & 0.531 & 0.893 & 0.932 & 0.947 \Bstrut\\
			\hline 
			Monodepth2 \cite{godard2019digging} & MS & Eigen Split & KF & 0.082 
			& 0.405 & 3.129 & 0.127 & 0.931 & 0.985 & \underline{0.996} 
			\Tstrut\\
			PackNet \cite{guizilini20203d} & MS & CS+Eigen Split & KF & 0.080 
			& 0.331 & 2.914 & 0.124 & 0.929 & 0.987 & \textbf{0.997} \\
			PackNet \cite{guizilini20203d} & MS, D & CS+Eigen Split & KF & 
			0.077 & \textbf{0.290} & 2.688 & 0.118 & 0.935 & 0.988 & 
			\textbf{0.997} \\
			DORN \cite{fu2018deep} & D & Eigen Split & KF & 0.077 & 
			\textbf{0.290} & 2.723 & 0.113 & 0.949 & 0.988 & \underline{0.996} 
			\Bstrut\\
			\hline
			DeepMVS \cite{huang2018deepmvs} & D & Odom. Split & KF+2 & 0.103 & 1.160 & 3.968 & 0.166 & 0.896 & 0.947 & 0.978 \Tstrut\\
			DeepMVS \cite{huang2018deepmvs} (pretr.) & D & Odom. Split & KF+2 & 0.088 & 0.644 & 3.191 & 0.146 & 0.914 & 0.955 & 0.982 \\
			DeepTAM \cite{zhou2018deeptam} (only FB) & MS, D* & Odom. Split & KF+2 & 0.059 & 0.474 & 2.769 & 0.096 & 0.964 & 0.987 & 0.994 \\
			DeepTAM \cite{zhou2018deeptam} (1x Ref.) & MS, D* & Odom. Split & 
			KF+2 & \underline{0.053} & 0.351 & \underline{2.480} & 
			\underline{0.089} & 
			\underline{0.971} & \underline{0.990} & 
			0.995 \Bstrut \\
			\hline
			\textbf{MonoRec} & MS, D* & Odom. Split & KF+2 & \textbf{0.050} & 
			\underline{0.295} & \textbf{2.266} & \textbf{0.082} & 
			\textbf{0.973} & \textbf{0.991} 
			& \underline{0.996} 
			\Tstrut \\
			\hline
		\end{tabular}\\[-0.5cm]
		\caption{\textbf{Quantitative Results on KITTI}: Comparison between 
		MonoRec and other methods on our KITTI test set. 
		The Dataset column shows the training dataset used by the corresponding 
		method and please note that Eigen split is a \textit{superset} of our 
		odometry split.
		Best / Second best results are marked 
		\textbf{bold} / \underline{underlined}. The evaluation result shows 
		that our 
		method achieves overall the best performance.
		\textbf{Legend}: M: \textit{Monocular images}, S: \textit{Stereo 
		images}, D: \textit{GT depth}, D*: \textit{Depths from DVSO}, KF: 
		\textit{Keyframe}, KF + 2: \textit{Keyframe + 2 mono frames}, CS: 
		\textit{Cityscapes} \cite{cordts2016cityscapes}, pretr.: 
		\textit{Pretrained 
		network}, FB: \textit{Fixed band module of DeepTAM}, Ref.: 
		\textit{Narrow band refinement 
		module of DeepTAM}}
	    \label{tab:quant_kitti}
	\end{center}
\end{table*}

To evaluate the proposed method, we first compare against state-of-the-art 
monocular depth prediction and \ac{mvs} methods with our train/test split of 
the 
KITTI dataset~\cite{geiger2012we}. Then, we perform extensive ablation studies 
to 
show 
the efficacy of our design choices. In the end, we demonstrate the 
generalization capabilities of different methods on Oxford 
RobotCar~\cite{Maddern2017robotcar} and TUM-Mono~\cite{engel2016monodataset} 
using the model trained on KITTI.

\begin{figure}
	\begin{center}
		\vspace{-0.5cm}
		\begin{subfigure}{0.45\textwidth}
			\includegraphics[width=\textwidth]{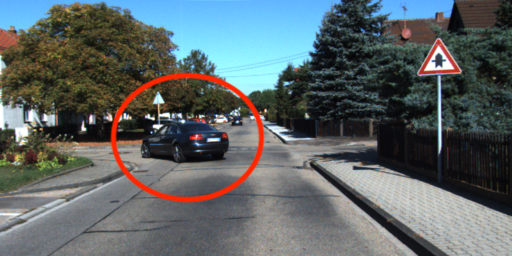}
			\vspace{-.5cm}
			\caption{Keyframe}
		\end{subfigure}
		\begin{subfigure}{0.45\textwidth}
			\includegraphics[width=\textwidth]{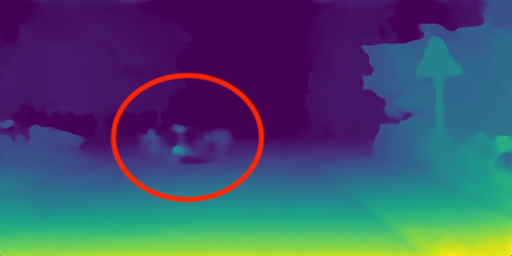}
			\vspace{-.5cm}
			\caption{W/o MaskModule}
		\end{subfigure}
		\begin{subfigure}{0.45\textwidth}
			\includegraphics[width=\textwidth]{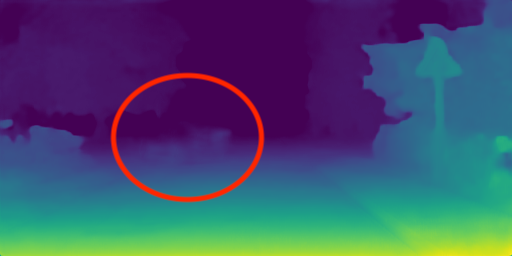}
			\vspace{-.5cm}
			\caption{MaskModule}
		\end{subfigure}
		\begin{subfigure}{0.45\textwidth}
			\includegraphics[width=\textwidth]{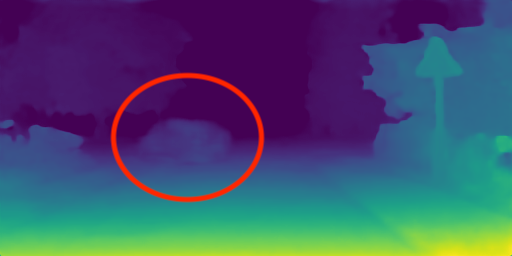}
			\vspace{-.5cm}
			\caption{MaskModule+D.Ref.}
		\end{subfigure}\\[-0.5cm]
		\caption{\textbf{Qualitative Improvement}: Effects of cost volume masking 
		and depth refinement.\vspace{-.1cm}}
		\label{fig:depth_pred_abl}
	\end{center}
\end{figure}

\subsection{The KITTI Dataset}

The Eigen split~\cite{eigen2015predicting} is the most popular training/test 
split for evaluating depth estimation on KITTI. We cannot make use of 
it directly since MonoRec requires temporally continuous 
images with estimated poses. Hence, we select our training/testing 
splits as the intersection between the KITTI Odometry benchmark and 
the Eigen split, which results in 13714/8634 
samples for training/testing. We obtain the relative poses 
between the images from the
monocular \ac{vo} system DVSO~\cite{yang2018deep}. During training, we also 
leverage the point clouds generated by DVSO as the sparse depth supervision 
signals. For training MaskModule we only use images 
that contain moving objects in the generated auxiliary masks, 2412 in total. For all the following evaluation results we use the improved ground 
truth~\cite{uhrig2017sparsity} and cap depths at \SI{80}{m}.

We first compare our method against the recent state of the art including an optimization based method (Colmap),
self-supervised monocular methods 
(MonoDepth2 and PackNet), a semi-supervised monocular method using sparse LiDAR 
data (PackNet), a supervised monocular method (DORN) and \ac{mvs} methods 
(DeepMVS and DeepTAM), shown in Table~\ref{tab:quant_kitti}. Note that 
the training code of DeepTAM was not published, we therefore implemented it 
ourselves for training and testing using our split to deliver a fair 
comparison. Our 
method outperforms all the other methods with a notable margin despite relying 
on images only without using LiDAR ground truth for training.

\begin{figure*}
	\begin{center}
		\vspace{-0.4cm}
		\includegraphics[width=\textwidth]{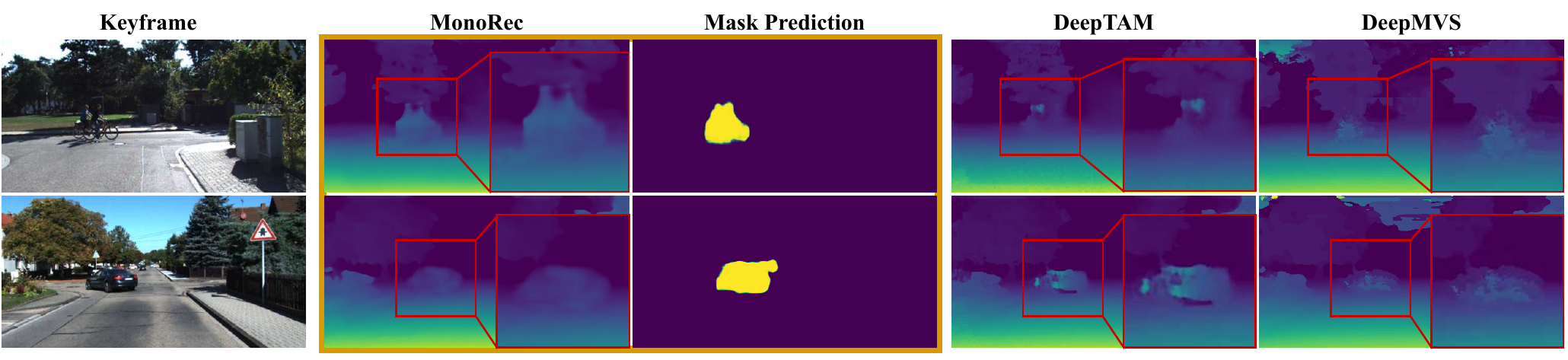}\\[-0.5cm]
		\caption{\textbf{Comparison on Moving Objects Depth Estimation}: In 
			comparison to other \acs{mvs} methods, MonoRec is able to predict 
			plausible depths. Furthermore, the depth 
			prediction has less noise and artifacts in static regions of the 
			scene.\vspace{-.3cm}
		}
		\label{fig:moving_objects}
	\end{center}
\end{figure*}

This is also clearly reflected in the qualitative results shown in 
Figure~\ref{fig:qualitative_results}. Compared with monocular depth estimation 
methods, our method delivers very sharp edges in the depth maps and can recover 
finer details. In comparison to the other \ac{mvs} methods, it can better deal 
with moving objects, which is further illustrated in 
Figure~\ref{fig:moving_objects}.

A single depth map usually cannot really reflect the quality for large scale 
reconstruction. We therefore also visualize the accumulated points using the 
depth maps from multiple frames in lower part of 
Figure~\ref{fig:qualitative_results}. We can 
see that our method can deliver very high quality reconstruction and, due to 
our 
MaskModule, is able to remove artifacts caused by moving objects. We urge 
readers to watch the supplementary video for more convincing comparisons.

\begin{table*}
	\begin{center}
			\scriptsize
			\begin{tabular}{|l|cccc|cccc|ccc|}
				\hline
				Model & SSIM & MaskModule & D. Ref. & M. Ref. & 
				\cellcolor{drawioOrange}Abs Rel & \cellcolor{drawioOrange}Sq 
				Rel & \cellcolor{drawioOrange}RMSE & 
				\cellcolor{drawioOrange}RMSE$_{log}$ & 
				\cellcolor{drawioGreen}$\delta<1.25$ & 
				\cellcolor{drawioGreen}$\delta<1.25^2$ & 
				\cellcolor{drawioGreen}$\delta<1.25^3$ \Tstrut\\
				\hhline{============}
				Baseline &  &  &  & & 0.056 & 0.342 & 2.624 & 0.092 & 0.965 & 0.990 & 0.994 \Tstrut\\
				Baseline & \checkmark &  &  &  & 0.054 & 0.346 & 2.444 & 0.088 
				& 0.970 & 0.989 & \underline{0.995} \Bstrut\\
				\hline
				MonoRec & \checkmark & \checkmark &  & \checkmark & 
				0.054 & 0.306 & 2.372 & 0.087 & 0.970 & \underline{0.990} & 
				\underline{0.995} \Tstrut\\
				MonoRec & \checkmark &  & \checkmark &  & 
				\underline{0.051} & 0.346 
				& 2.361 & \underline{0.085} & \underline{0.972} & 
				\underline{0.990} & \underline{0.995} 
				\\
				MonoRec & \checkmark & \checkmark & \checkmark &  & 
				0.052 & \underline{0.302} & \underline{2.303} & 0.087 & 0.969 & 
				\underline{0.990} & 
				\underline{0.995} \\
				\textbf{MonoRec} & \checkmark & \checkmark & \checkmark & 
				\checkmark & \textbf{0.050} & \textbf{0.295} & \textbf{2.266} & 
				\textbf{0.082} & 
				\textbf{0.973} & \textbf{0.991} & \textbf{0.996} \\
				\hline
		\end{tabular}\\[-0.5cm]
			\caption{\textbf{Ablation Study}: 
			Baseline consists of only DepthModule using the unmasked cost 
			volume (CV). Baseline without SSIM uses a 5x5 patch that has same 
			receptive field as SSIM. Using 
			SSIM to 
			form CV gives a significant improvement.
			For MonoRec, only the addition of MaskModule
			without refinement does not yield significant improvements. 
			The DepthModule refinement gives a major 
			improvement. The best performance is achieved by combining all the 
			proposed components.} \label{tab:ablation_depth}
	\end{center}
\end{table*}

\noindent \textbf{Ablation Studies}. We also investigated the contribution of the 
different components towards the method's performance.
Table~\ref{tab:ablation_depth} shows quantitative results of our ablation studies, which confirm that all our proposed contributions improve the depth prediction over the baseline method.
Furthermore, Figure~\ref{fig:depth_pred_abl} demonstrates the qualitative
improvement achieved by MaskModule and refinement training.

\subsection{Oxford RobotCar and TUM-Mono}

\begin{figure}[H]
	\begin{center}
		\includegraphics[width=.95\textwidth]{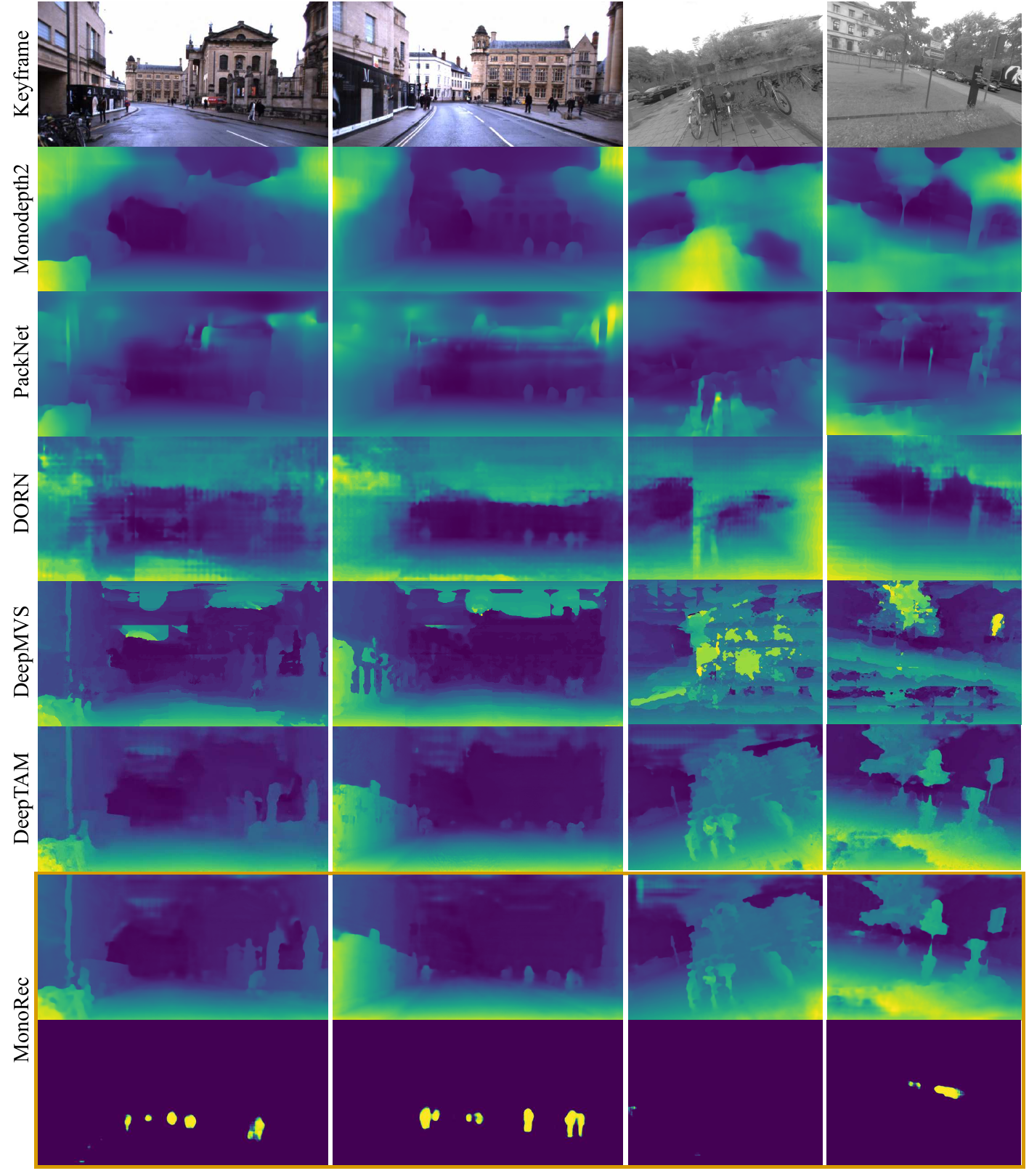}\\[-0.5cm]	
		\caption{\textbf{Oxford RobotCar and TUM-Mono}: All results 
			are obtained by the respective best-performing variant in 
			Table~\ref{tab:quant_kitti}. MonoRec shows stronger generalization 
			capability than the monocular methods. Compared to DeepMVS and 
			DeepTAM, MonoRec delivers depth maps with less artifacts and 
			predicts the moving object masks in addition.\vspace{-1cm}}
		\label{fig:oxtumrc}
	\end{center}
\end{figure}

To demonstrate the generalization capabilities of MonoRec, we test 
our KITTI model on the Oxford RobotCar dataset and the TUM-Mono dataset. 
Oxford RobotCar is a street view dataset and shows a similar motion pattern and view 
perspective to KITTI. TUM-Mono, however, is recorded by a handheld 
monochrome camera, so it demonstrates very different motion and image quality 
compared to KITTI. The results are shown in 
Figure~\ref{fig:oxtumrc}. The monocular methods struggle to 
generalize to a new context.
The compared 
\ac{mvs} methods show more artifacts and cannot predict 
plausible depths for the moving objects. In contrast our method is able to 
generalize well to the new scenes for both depth and moving object 
predictions.
Since Oxford RobotCar also provides LiDAR depth data, we further show a 
quantitative evaluation in the supplementary material.

\section{Conclusion}
\label{sec:conclusion}

We have presented MonoRec, a deep architecture that estimates accurate dense \ac{3d} 
reconstructions from only a single moving camera. We first propose to use 
\ac{ssim} as the photometric measurement to construct the cost volumes. To 
deal with 
dynamic objects, we propose a novel 
MaskModule which predicts moving object masks from the input cost volumes. With 
the predicted masks, the proposed DepthModule is able to estimate accurate 
depths for both static and dynamic objects. Additionally, we propose a novel 
multi-stage training scheme together with a semi-supervised loss formulation 
for training the depth prediction. All combined, 
MonoRec is able to outperform the state-of-the-art \ac{mvs} and monocular depth 
prediction methods both qualitatively and quantitatively on KITTI and also 
shows strong generalization capability on Oxford RobotCar and TUM-Mono. We 
believe that this capacity to recover accurate dense 3D reconstructions from a 
single moving camera will help to establish the camera as the lead sensor for 
autonomous systems.
\vfill\eject

{\footnotesize
\paragraph{Acknowledgement}
This work was supported by the Munich Center for Machine Learning and by the ERC Advanced Grant SIMULACRON.}

{\small
	\bibliographystyle{ieee_fullname}
	\bibliography{bibliography}
}

\clearpage
\appendix
\renewcommand{\thesection}{\Alph{section}}
{\noindent\Large\textbf{Supplementary Material}}

\section{Introduction}
In this supplementary material, we provide additional details in extension to our main paper.
This mainly includes more implementation details (Sec.~\ref{sec:impl}) 
and additional experimental results (Sec.~\ref{sec:exp}).

\section{Implementation Details}\label{sec:impl}
\begin{figure*}
	\begin{center}
		\includegraphics[width=\textwidth]{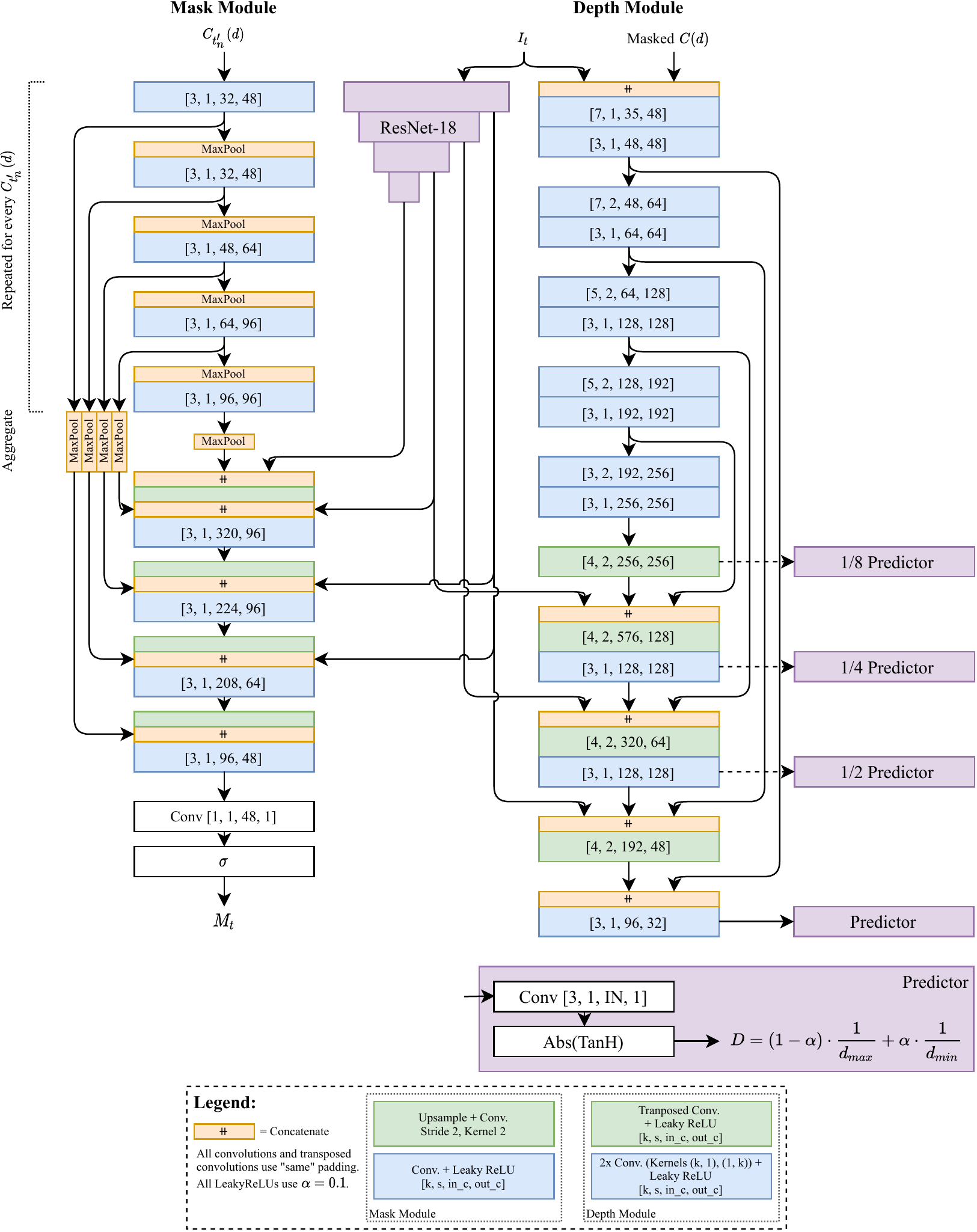}
	\end{center}
	\vspace{-.25cm}
	\caption{\textbf{Detailed Architecture of MonoRec}}
	\label{fig:detail_arc}
\end{figure*}

The exact details of our network architecture can be observed in Figure~\ref{fig:detail_arc}.

As described in section 3.4 of the main paper, we use several different error 
thresholds to generate the auxiliary training masks. Since for this task it is 
more important for the error metric to be semantically consistent instead of 
very detailed, we use perceptual error instead of absolute differences or SSIM. 
To this end, we employ the first 9 layers of a pretrained VGG-16 network from 
the PyTorch model zoo. The per-pixel error between two images is defined as the 
mean squared error between the respective feature vectors for the respective pixels. 
The thresholds are as follows: 
(1) $pe_{t^S}^t (\mathbf{x}, D_t(\mathbf{x})) > 12$ 
(2) $\overline{pe_{t'}^t} (\mathbf{x}, D_t^{S}(\mathbf{x})) > 8$
(3) $\max\{\frac{D_t(\mathbf{x})}{D_t^{S}(\mathbf{x})}, \frac{D_t^{S}(\mathbf{x})}{D_t(\mathbf{x})}\} > 1.5$.
If at least two out of these conditions are fulfilled a pixel is considered to be moving. 
To ensure temporal consistency of the moving object masks, we match every detected 
segmentation mask with masks from the previous and the following frame. The matched 
segmentation masks have to be from the same object class and have a minimum IoU of $0.25$. 
A segmentation mask is accepted as a moving object, if it itself and the matched 
segmentation masks contain on average more than $40\%$ moving pixels.

\section{Additional Experiments}\label{sec:exp}
We provide additional experimental results.
This comprises more extensive ablation studies (Sec.~\ref{sec:ablation_studies}) where we specifically evaluate the performance of the MaskModule.
Furthermore, the effect of different model configurations is evaluated.

We also provide some of the failure cases in which our method does not achieve 
optimal performance (Sec.~\ref{sec:failure_cases}).

In addition to the qualitative generalization capabilities of our method 
presented in the main paper, we also provide quantitative results obtained from 
the Oxford RobotCar dataset~\cite{Maddern2017robotcar} 
(Sec.~\ref{sec:oxford_robotcar}) and the TUM RGB-D 
dataset~\cite{sturm12iros} (Sec.~\ref{ssec:tum_rgbd}).

In Sec.~\ref{ssec:further_quant}, we show the quantitative evaluation against 
two other monocular dense reconstruction methods in dynamic 
scenes~\cite{ranftl2016dense,russell2014video}.

\subsection{Ablation Studies}\label{sec:ablation_studies}

In the ablation studies presented in the main paper, we focused on the overall performance on \acused{mvs}\ac{mvs} depth prediction and the contribution of the different components.
Here, we pay attention to the MaskModule and its performance with respect to masking out dynamic objects.
Furthermore, we evaluation different model configurations.  

\begin{table}
	\begin{center}
		\scriptsize
		\begin{tabular}{|l|ccc|}
			\hline
			Model & \cellcolor{drawioGreen}Prec & \cellcolor{drawioGreen}Rec & \cellcolor{drawioGreen}IoU  \Tstrut\\
			\hhline{====}
			Baseline (only ResNet) & 0.017 & 0.658 & 0.016 \Tstrut\\
			Baseline (only cost volume) & 0.230 & 0.642 & 0.204 \Bstrut\\
			\hline
			Baseline & 0.260 & 0.678 & 0.232 \Tstrut\\
			\textbf{Mask Refinement} & \textbf{0.374} & \textbf{0.748} & \textbf{0.300}\\
			\hline
		\end{tabular}\\[-0.5cm]
		\caption{\textbf{Ablation Study - MaskModule}: Results for the masks predicted by our MaskModule compared to the auxiliary masks on the proposed KITTI Odometry \cite{geiger2012we} test set using different versions of our model. \textbf{Note}: The auxiliary masks can not be compared to ground truth as they themselves contain many mistakes (both missed detections and miss-classifications). Our \textbf{Baseline} model was only trained with the auxiliary masks. \textbf{Mask Refinement} describes our model after the mask refinement training. It improves the performance across all metrics.} \label{tab:ablation_mask}
	\end{center}
\end{table}

\begin{table*}
	\begin{center}
			\scriptsize
			\begin{tabular}{|c|l|cccc|ccc|}
				\hline
				& Model & \cellcolor{drawioOrange}Abs Rel & \cellcolor{drawioOrange}Sq Rel & \cellcolor{drawioOrange}RMSE & \cellcolor{drawioOrange}RMSE$_{log}$ & \cellcolor{drawioGreen}$\delta<1.25$ & \cellcolor{drawioGreen}$\delta<1.25^2$ & \cellcolor{drawioGreen}$\delta<1.25^3$ \Tstrut\\
				\hhline{=========}
				(a) & 4 Frames & 0.045 & 0.267 & 2.130 & 0.082 & 0.975 & 0.991 & 0.995 \Tstrut\\
				& 6 Frames & 0.046 & 0.271 & 2.163 & 0.087 & 0.972 & 0.989 & 0.995 \\
				& 320x640 & 0.052 & 0.309 & 2.230 & 0.084 & 0.970 & 0.990 & 0.995 \\
				& KITTI poses & 0.077 & 0.077 & 3.283 & 0.943 & 0.943 & 0.982 & 0.992 \\
				& MonoRec & 0.050 & 0.288 & 2.269 & 0.082 & 0.972 & 0.991 & 0.996 \Bstrut\\
				\hline
				(b) & M, D* Baseline & 0.059 & 0.494 & 2.764 & 0.096 & 0.966 & 0.987 & 0.994  \Tstrut\\
				& MS, D* Baseline & 0.054 & 0.346 & 2.444 & 0.088 & 0.970 & 0.989 & 0.995\\
				\hline
		\end{tabular}\\[-0.5cm]
		\caption{\textbf{Ablation Study - Model Configuration}: Depth prediction results using different model configurations. \textbf{(a)} All models use the same weights, that were trained with 2 frames, DVSO \cite{yang2018deep} poses and $256 \times 512$. \textbf{(b)} Mono vs.\ Mono + Stereo training of depth module.} \label{tab:ablation_config}		
	\end{center}
\end{table*}

\begin{figure}
	\begin{center}
		\includegraphics[width=\textwidth]{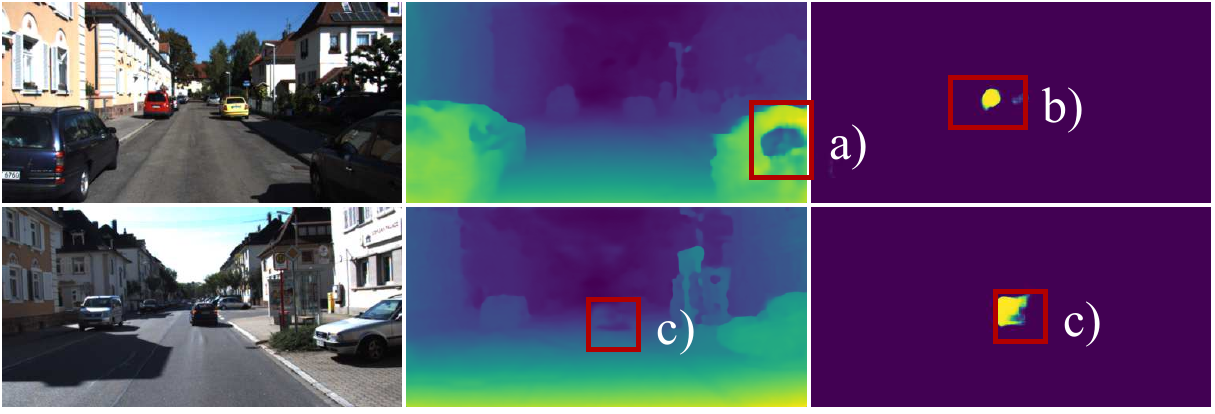}
	\end{center}
	\caption{\textbf{Failure Cases}: a)\ Non-lambertian surfaces, 
	especially ones that are very close, can lead to mis-predictions 
	due to a wrong cost volume prior. b)\ The MaskModule sometimes detects 
	the focal point, if far away, as a moving object. The effect is 
	minimal, because these pixels are not used for reconstruction. 
	c) If the predicted mask does not cover the moving object entirely, the 
	network might produce artifacts due wrong cost volume priors.}
	\label{fig:failure}
\end{figure}

\subsubsection{MaskModule}
For MaskModule it is more important to filter out all moving objects reliably than having a very high precision, since DepthModule is able to fill out small missing patches in the cost volume.
Therefore, in the trade-off between recall and precision we put higher emphasis on recall.
As baseline we consider MaskModule only trained based on the the auxiliary masks.
This baseline is compared against the mask prediction after refinement training.
The baseline already achieves fairly high recall, however, the precision is not very strong (see Table~\ref{tab:ablation_mask}).
Through the refinement training, which puts the mask prediction into direct context with the cost volume input, the performance is improved across all metrics, especially the precision.

\subsubsection{Model Configuration} The standard configuration of our model receives a keyframe and two additional mono frames (the one before and after the keyframe) at a resolution $256 \times 512$ as well as poses generated by DVSO~\cite{yang2018deep} as input.
However, our implementation is very flexible.
It can take any number of frames at any resolution that is a multiple of 16.
Furthermore, the pose source can easily be replaced, e.g. by another \ac{vo} algorithm or other sensors (e.g. INS).
The results in Table~\ref{tab:ablation_config} shows that by feeding more frames into the model, one can, in fact, improve the performance.
However, this effect saturates after a certain number of frames.
Interestingly, our model works significantly worse with the ground truth poses provided by KITTI Odometry \cite{geiger2012we}.
We believe that this is because DVSO \cite{yang2018deep} computes poses solely based on monocular photometric error, similarly to the way our cost volume is built.
Furthermore, since the ground truth poses in KITTI are obtained based on an INS system, they might be locally less accurate than the \ac{vo} poses and not perfectly synchronized with the images.
Finally, our model does not seem to significantly benefit from a larger image input size.

\begin{table}
	\begin{center}
		\scriptsize
		\begin{tabular}{|l|cc|c|}
			\hline
			Method & \cellcolor{drawioOrange}Abs Rel & \cellcolor{drawioOrange}RMSE & \cellcolor{drawioGreen}$\delta<1.25$ \Tstrut\\
			\hhline{====}
			Monodepth2 \cite{godard2019digging} & 0.220 & 7.328 & 0.616 \Tstrut\\
			PackNet \cite{guizilini20203d} & 0.233 & 7.512 & 0.606 \\
			PackNet \cite{guizilini20203d}(supervi.) & 0.229 & 7.983 & 0.620 \\
			DORN \cite{fu2018deep} & 0.215 & 7.966 & 0.651 \Bstrut\\
			\hline
			DeepMVS \cite{huang2018deepmvs}& \textbf{0.142} & 7.379 & \underline{0.780} \Tstrut\\
			DeepMVS \cite{huang2018deepmvs} (pretr.) & 0.153 & \textbf{6.656} & 0.770 \\
			DeepTAM \cite{zhou2018deeptam} (only FB) & 0.154 & 7.355 & 0.776 \\
			DeepTAM \cite{zhou2018deeptam} (1x Ref.) & 0.152 & 7.211 & 0.749 \Bstrut\\
			\hline
			MonoRec & \underline{0.143} & \underline{7.180} & \textbf{0.806} \Tstrut\\
			\hline
		\end{tabular}\\[-0.5cm]
		\caption{\textbf{Oxford RobotCar}: Quantitative performance of different models on the Oxford RobotCar dataset. Best / Second best results are marked \textbf{bold} / \underline{underlined}.} \label{tab:oxrc}
	\end{center}
\end{table}

\subsection{Failure Cases}\label{sec:failure_cases}
In Figure~\ref{fig:failure} we visualize typical failure cases of our method. Some of the show failure cases, like the ones caused by non-lambertian surfaces are typical for \ac{mvs} methods. Other failures are a result of miss-detections of the MaskModule. However, at least partially, those miss-detections can be compensated by our DepthModule.

\subsection{Oxford RobotCar Dataset}\label{sec:oxford_robotcar}

In Table~\ref{tab:oxrc} we show the quantitative results of Oxford RobotCar 
generated with the official long sample sequence.
To get the ground truth, we aggregated multiple LiDAR scans within a range of 
\SI{0.25}{s} before and 
after the frame timestamp and transformed it using the odometry poses.
Note that, due to the short sequence and the low quality of LiDAR data, one has 
to consider the provided numbers with caution. Nevertheless, considering the 
numbers our method performs arguably overall the best among all evaluated 
methods.

\subsection{TUM RGB-D}\label{ssec:tum_rgbd}

To further demonstrate MonoRec's generalization capabilities, we also performed 
quantitative analysis on the indoor TUM RGB-D \cite{sturm12iros} dataset using 
the models trained on KITTI. Table~\ref{tab:tumrgbd} shows that MonoRec 
delivers better results compared to other methods.

\begin{table}
	\begin{center}
		\scriptsize
		\begin{tabular}{|l|cc|c|}
			\hline
			Method & \cellcolor{drawioOrange}Abs Rel & \cellcolor{drawioOrange}RMSE & \cellcolor{drawioGreen}$\delta<1.25$ \Tstrut\\
			\hhline{====}
			MonoDepth2 \cite{godard2019digging} & 0.353 & 1.240 & 0.458 \Tstrut\\
			DeepTAM \cite{zhou2018deeptam} (1xRef) & \textit{0.210} & \textit{0.792} & \textit{0.701} \\
			MonoRec & \textbf{0.189} & \textbf{0.756} & \textbf{0.725} \Bstrut\\
			\hline
		\end{tabular}\\[-0.5cm]
		\caption{\textbf{TUM RGB-D}: Quantitative performance of different 
		methods on the TUM RGB-D dataset. Specifically, we evaluate on the 
		{\scriptsize 
		\texttt{freiburg3\_long\_office\_household}} sequence. 
		Best / Second best results are marked \textbf{bold} / 
		\underline{underlined}. All methods are trained on KITTI and 
		MonoRec shows stronger generalization capability.} \label{tab:tumrgbd}
	\end{center}
\end{table}

\subsection{Further Quantitative Evaluations}\label{ssec:further_quant}

In Table~\ref{tab:supp_quant} we show quantitative comparisons to DenseMono 
\cite{ranftl2016dense} and VideoPopup \cite{russell2014video}. These methods, 
like MonoRec, aim to deliver accurate depths for dynamic scenes and make use of 
consecutive frames as input additional to the keyframe. Both methods employ 
classical optimization methods instead of neural networks. The evaluation 
results suggest that MonoRec performs better than DenseMono and VideoPopup.

\begin{table}
	\begin{center}
		\scriptsize
		\begin{tabular}{|l|cc|c|}
			\hline
			Method & \cellcolor{drawioOrange}Abs Rel & \cellcolor{drawioOrange}RMSE & \cellcolor{drawioGreen}$\delta<1.25\Tstrut$\\
			\hhline{====}
			DenseMono \cite{ranftl2016dense} & 0.148 & 2.408 & not provided \Tstrut\\
			MonoRec & \textbf{0.079} & \textbf{1.469} & 0.949 \Bstrut\\
			\hhline{----}
			VideoPopup \cite{russell2014video} & 0.154 & 2.631 & 0.752 \Tstrut\\
			MonoRec & \textbf{0.054} & \textbf{2.304} & \textbf{0.970} \Bstrut\\
			\hline
		\end{tabular}\\[-0.5cm]
		\caption{\textbf{Quantitative Results - Further Methods}: Comparisons of depth evaluation to further methods. Best results are marked \textbf{bold}. In the comparison to DenseMono \cite{ranftl2016dense}, sequences 11-21 of the KITTI odometry dataset are used. For the comparison to VideoPopup, sequence 05 of the KITTI odometry dataset is used.} \label{tab:supp_quant}
	\end{center}
\end{table}

\end{document}